\documentclass[11pt]{article}
\usepackage[utf8]{inputenc}
\usepackage[parfill]{parskip}    
\usepackage{graphicx}
\usepackage{amssymb, amsmath, amsthm}
\usepackage{epstopdf}
\usepackage{mathrsfs}
\usepackage[marginratio=1:1,height=600pt,width=440pt,tmargin=60pt]{geometry}
\usepackage{layout, color}
\usepackage{bbm}

\usepackage{authblk}

\usepackage{algorithm}
\usepackage{algpseudocode}

\usepackage{setspace}
\usepackage{mathtools}

\usepackage{hyperref}
\usepackage{url}

\usepackage{times}
\usepackage{epsfig}
\usepackage{graphicx}
\usepackage{amsmath}
\usepackage{amssymb}
\usepackage{subfigure}
\usepackage{mathrsfs}
\usepackage{booktabs}
\usepackage{multirow}
\usepackage{array}

\usepackage{mathrsfs,amsthm,amsmath,amssymb}

\newcommand{\PreserveBackslash}[1]{\let\temp=\\#1\let\\=\temp}
\newcolumntype{C}[1]{>{\PreserveBackslash\centering}p{#1}}
\newcolumntype{R}[1]{>{\PreserveBackslash\raggedleft}p{#1}}
\newcolumntype{L}[1]{>{\PreserveBackslash\raggedright}p{#1}}

\usepackage{comment}
\usepackage{mdwlist}

\usepackage{caption}

\usepackage{verbatim}
\usepackage{algorithm}  
\usepackage{algpseudocode}

\usepackage{wrapfig,lipsum,booktabs}

\usepackage{enumitem}
\usepackage{xcolor}
\newif\ifgooditem
\gooditemtrue

\theoremstyle{plain}

\usepackage{amsmath}

\usepackage{multirow}

\usepackage{algorithm}  
\usepackage{algpseudocode}

\newcommand{\atom}{\mathbf{D}}
\newcommand{\coef}{\mathbf{A}}
\newcommand{\inp}{\mathbf{X}}
\newcommand{\oup}{\mathbf{Y}}

\newcommand{\K}{\mathbf{K}}
\newcommand{\itr}{\mathbf{Z}}
\newcommand{\WP}{\mathbf{P}}
\newcommand{\WQ}{\mathbf{Q}}
\newcommand{\MR}{\multirow}
\title{ACDC: Weight Sharing in Atom-Coefficient Decomposed Convolution}

\author[1]{Ze Wang}
\author[2]{Xiuyuan Cheng}
\author[2]{Guillermo Sapiro}
\author[1]{Qiang Qiu}

\affil[1]{Purdue University \authorcr {\tt\small \{zewang, qqiu\}@purdue.edu}}
\affil[2]{Duke University \authorcr {\tt\small \{xiuyuan.cheng, guillermo.sapiro\}@duke.edu}}

\date{}

\begin{document}
	
\maketitle

\begin{abstract}
	
	Convolutional Neural Networks (CNNs) are known to be significantly over-parametrized, and difficult to interpret, train and adapt. 
	In this paper, we introduce a structural regularization across convolutional kernels in a CNN. In our approach, each convolution kernel is first decomposed as 2D dictionary atoms linearly combined by coefficients. 
	The widely observed correlation and redundancy in a CNN hint a common low-rank structure among the decomposed coefficients, which is here further supported by our empirical observations. 
	We then explicitly regularize CNN kernels by enforcing decomposed coefficients to be shared across sub-structures, while leaving each sub-structure only its own dictionary atoms, a few hundreds of parameters typically, which leads to dramatic model reductions. 
	We explore models with sharing across different sub-structures to cover a wide range of trade-offs between parameter reduction and expressiveness. 
	Our proposed regularized network structures open the door to better interpreting, training and adapting deep models. 
	We validate the flexibility and compatibility of our method by image classification experiments on multiple datasets and underlying network structures, and show that CNNs now maintain performance with dramatic reduction in parameters and computations, e.g., only 5\% parameters are used in a ResNet-18 to achieve comparable performance. Further experiments on few-shot classification show that faster and more robust task adaptation is obtained in comparison with models with standard convolutions.

\end{abstract}

\section{Introduction}

Convolutional Neural Networks (CNNs) have achieved remarkable progresses on solving challenging tasks. 
The successes stimulate research directions that further improve CNNs from various angles, including network structures \cite{he2016deep,howard2017mobilenets,ma2018shufflenet,simonyan2014very,zagoruyko2016wide,zhang2018shufflenet}, fast adaptations \cite{finn2017model,lopez2017gradient,shin2017continual}, parameter efficiency \cite{cheng2017survey,han2015deep,luo2017thinet,savarese2019learning}, and interpretability \cite{selvaraju2017grad,zhou2016learning}.
With the trend of deeper and wider network structures with hundreds of millions of parameters, such investigations become even more pressing.
The aforementioned challenges can be partially attributed to the under-regularized structures of convolutional kernels in a CNN, which are typically of very high dimensions and trained independently from random initializations.
While recent works on efficient convolution operations \cite{Chollet_2017_CVPR,howard2017mobilenets} alleviate the long recognized over-parametrization problem of deep CNNs, kernels across different convolutional layers are still modeled as isolated and independent groups of parameters, among which interactions only happen during feature and gradient propagations.
Modeling kernels by shared structures has been empirically studied \cite{ha2016hypernetworks,savarese2019learning}, which sheds the light on explicitly modeling the underlying common structures across kernels, and confirms the widely observed redundancies in deep network parameters \cite{michel2019sixteen,raghu2017svcca}. Studies on deep representations \cite{kornblith2019similarity,morcos2018insights,raghu2017svcca} suggest that, under certain linear transforms, deep features across layers are actually highly correlated. 
Such observations, together with the well recognized redundancies in parameters, motivate us to further exploit such correlation to enforce explicit structural regularizations over kernels. The work here presented provides a foundational plug-and-play framework to introduce structure in convolution kernels via coefficient sharing within and between layers, resulting in significantly smaller and more interpretable networks with maintained performance, even obtaining performance improvement for some tasks.

We first perform atom-coefficient decompositions to convolution kernels, in which each kernel is decomposed as 2D dictionary atoms linearly combined by coefficients. 
A standard convolution layer can now be decomposed into two: a dictionary atom sub-layer involving spatial-only convolution with the dictionary atoms, followed by a coefficient sub-layer that linearly combines feature channels from the atom layer. 
Due to the underlying cross-layer correlations, after we properly align the outputs of both sub-layers across the network's multiple layers through canonical correlation analysis (CCA), we obtain a low rank structure for those dictionary coefficients. 
This observation hints us to enforce shared coefficients across sub-structures, e.g., layers. 
By sharing coefficients, each sub-structure is now left with only dictionary atoms, which typically include only a few hundreds of parameters and lead to dramatic model reduction.
The focus of the paper is to introduce, derive, and fully explore such atom-coefficient decomposed convolution (ACDC) as a structural regularization to convolution kernels.
The easily constructed variants, e.g., with different numbers of dictionary atoms and coefficients sharing across different substructures, enable a wide coverage of trade-offs between parameter reduction and model expressiveness.
The explicitly regularized structures open the door to better interpreting, training, and adapting deep models. 

We perform extensive experiments on standard image classification datasets, and show that, by using variants of ACDC as plug-and-play replacements to the standard convolution in various off-the-shelf network architectures, different degrees of model reductions are achieved with comparable or even better accuracy. Some ACDC variants can substantially reduce the overall computation of a deep CNN. Further experiments on few-shot classification demonstrate the fast adaptation across tasks of the proposed method. 

Our main contributions are summarized as follows:
\begin{itemize}
	\item We introduce ACDC, a plug-and-play replacement to the standard convolution that achieves a structural regularization for kernels within a CNN for better interpretability, training, and adaptations.
	\item Highlighting the remarkable flexibility, we introduce variants of ACDC constructed easily by coefficient sharing within different network sub-structures and varying numbers of dictionary atoms.
	\item We validate the effectiveness of ACDC by plug-playing them into modern CNN architectures for various tasks.
\end{itemize}

\section{Atom-Coefficient Decomposed Convolution}
In this section, we start with a brief introduction of atom-coefficient decomposition motivated by dictionary learning and recent works on decomposed convolutional kernels. 
We then introduce the general idea of ACDC hinted by both the well recognized over-parametrization problem and the underlying cross-layer correlations of CNNs. Based on the idea of coefficients sharing enforced across network sub-structures, we describe in details how variants of ACDC are constructed as plug-and-play replacements to the standard convolution.

\subsection{Convolutional Kernel Decomposition}
\begin{figure}
	\resizebox{\textwidth}{!}{%
		\begin{minipage}{0.48\linewidth}
			\vspace{10mm}
			\includegraphics[width=\linewidth]{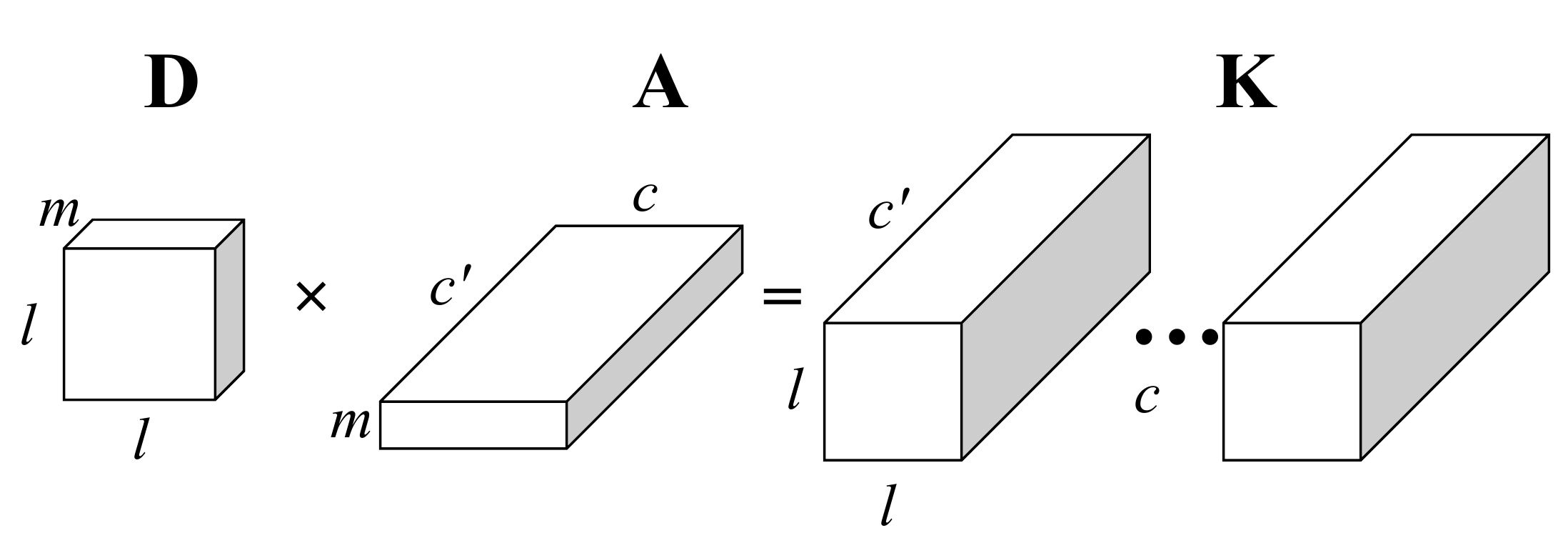}
			\vspace{5mm}
			\caption{Illustration of the atom-coefficient decomposition. A convolutional kernel $\K$ with $c\times c^\prime$ filters is reconstructed by multiplying $m$ 2D dictionary Atoms with sizes $l\times l$ and coefficients $\coef \in \mathbb{R}^{c \times c^\prime \times m}$.}
			\label{fig:dcf} 
		\end{minipage}
		\hspace{5mm}
		\begin{minipage}{0.5\linewidth}
			\includegraphics[width=\linewidth]{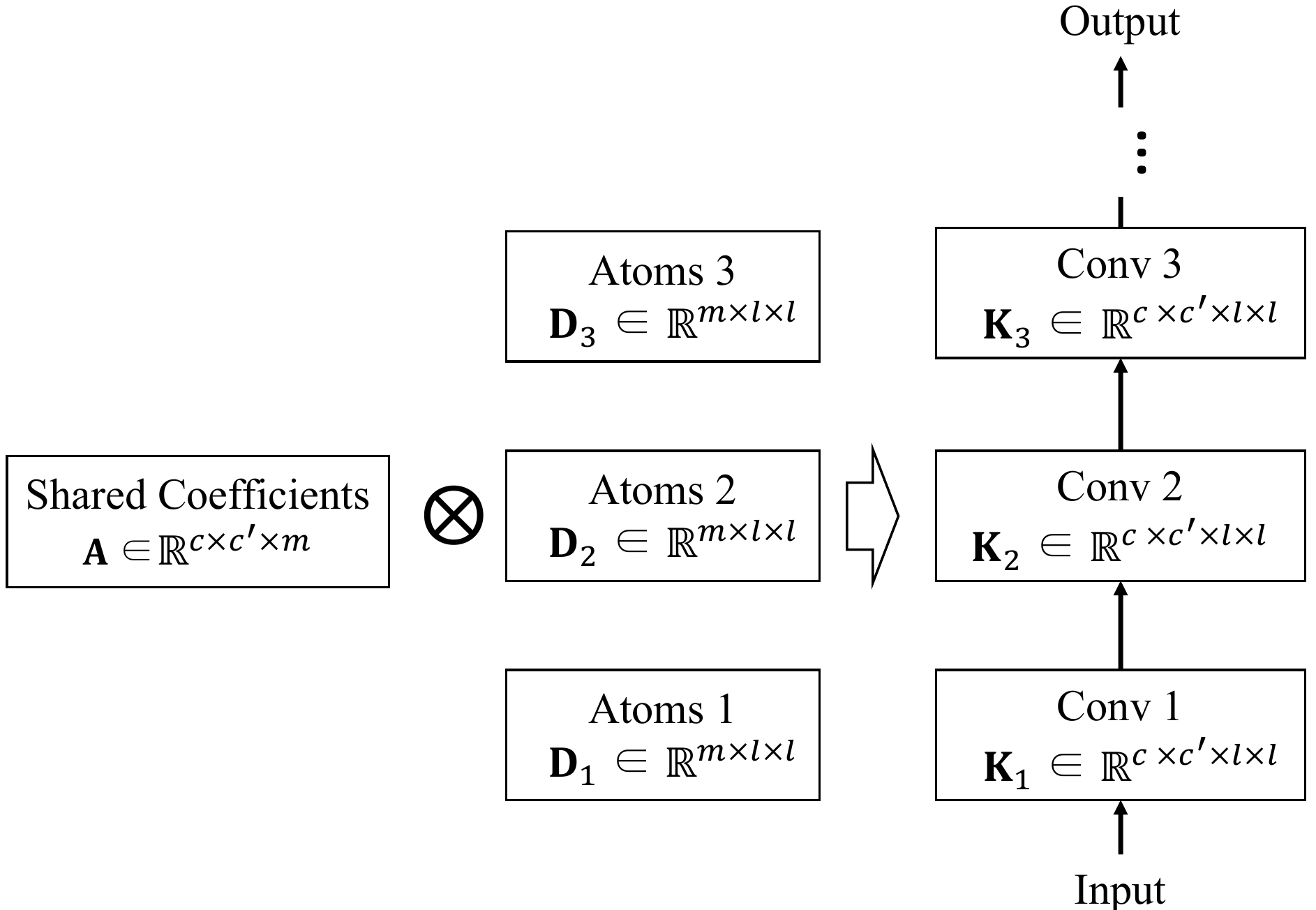}
			\caption{Illustration of \textit{ACDC-net} deployed in a simple CNN. $\otimes$ denotes matrix multiplication. Intermediate features are omitted. 
			}
			\label{fig:net} 
		\end{minipage}
	}
\end{figure}

Previous works have shown that a convolutional kernel in a CNN can be decomposed as a linear combination of pre-fixed basis \cite{qiu2018dcfnet}. 
In ACDC, we adopt a similar decomposition as shown in Figure~\ref{fig:dcf}, in which a convolutional kernel is represented as a linear combination of trainable 2D dictionary atoms.
After decomposition, a convolution layer with $c$-channel output $\oup$ and $c^\prime$-channel input $\inp$ becomes
\begin{equation}
\begin{aligned}
\label{eq:decomposed}
\oup = \K * \inp, 
\quad
\K = \atom \coef,
\end{aligned}
\end{equation}
where * denotes the convolution operation.
As illustrated in Figure~\ref{fig:dcf}, in (\ref{eq:decomposed}), a convolutional kernel $\K \in \mathbb{R}^{c \times c^\prime \times l \times l}$, which can be seen as a stack of $c \times c^\prime$ 2D convolutional filters with the size of $l \times l$, is reconstructed by multiplying $m$ 2D dictionary atoms $\atom \in \mathbb{R}^{m \times l \times l}$ with the corresponding linear coefficients $\coef \in \mathbb{R}^{c\times c^\prime m}$. Note that square kernels are assumed here for simplicity, while all kernel shapes are supported.
Since both convolution and tensor multiplication are commuting linear operations, a convolutional layer can now be decomposed into two: 
\begin{itemize}
	\item A dictionary \textit{atom sub-layer} where each atom involves spatial-only convolution with the dictionary atoms, i.e., $\itr \in \mathbb{R}^{c^\prime m \times h \times w} = \atom * \inp$; 
	\item A linear \textit{coefficient sub-layer} that linearly combines feature channels from the \textit{atom sub-layer}: $\oup \in \mathbb{R}^{c \times h \times w} = \coef \itr$. Note that $\itr$ here denotes \textit{atom sub-layer} outputs, and stride 1 and same padding are assumed for the sake of discussion.
\end{itemize}

\subsection{Correlation and Redundancy: The Motivation Behind}

Deep CNNs are long recognized to be over-parametrized. The very deep layers in modern CNN structures \cite{he2016deep,huang2017densely,zagoruyko2016wide} and the high-dimensional kernels with little structural regularizations lead to hundreds of millions of parameters. Such problem of over-parametrization is also observed in the studies of deep representations \cite{raghu2017svcca}, and empirically alleviated by new network structures \cite{Chollet_2017_CVPR,howard2017mobilenets}, network compressions, and parameter reduction methods \cite{ha2016hypernetworks,savarese2019learning}.

Meanwhile, recent studies on deep representations \cite{kornblith2019similarity,morcos2018insights,raghu2017svcca} have shown that there exists obvious correlations for features across layers within a CNN after proper linear alignments. 
Such observations are also supported by the success of deep network with residual learning \cite{he2016deep}, which explicitly formulates the layers as learning residual functions with reference to the layer inputs.
The correlation across features motivate us to explore and exploit correlations across kernels for structural regularizations.

We present here a motivating experiment pn MNIST by applying CCA alignments as in \cite{raghu2017svcca} to the \textit{atom sub-layer} outputs and the \textit{coefficient sub-layer} outputs of layer $i$ and layer $j$. Note that no parameter sharing is imposed here, and the network reports the same testing accuracy before and after kernel decomposition.
Formally, $c$, $m$, $d$, and $hw$ denote the number of channels, number of dictionary atoms, test set size, and the 2D feature dimensions, respectively.
The \textit{atom sub-layer} outputs of the $i$-th and $j$-th layer, $\itr_i$ and $\itr_j \in \mathbb{R}^{cm \times dhw}$, are firstly aligned by linear transformations $\WP_i$ and $\WP_j \in \mathbb{R}^{cm \times cm}$ that maximize the correlation $\rho_z = \max\limits_{\WP_i, \WP_j} corr(\WP_i \itr_i, \WP_j \itr_j)$. And similarly, the \textit{coefficient sub-layer} outputs of both layers, $\oup_i$ and $\oup_j \in \mathbb{R}^{c \times dhw}$, are aligned by $\WQ_i$ and $\WQ_j \in \mathbb{R}^{c \times c}$ that maximize the correlation $\rho = \max\limits_{\WQ_i, \WQ_j} corr(\WQ_i \oup_i, \WQ_j \oup_j)$. Omitting the layer indexes, the feed forwards of both layers can be rewritten as
\begin{equation}
\oup = \WQ \coef \WP^{-1} \WP (\atom * \inp).
\end{equation} 
By merging the transform into the coefficients $\coef$ by $\widetilde{\coef} = \WQ \coef \WP^{-1}$, we obtain `aligned coefficients' $\widetilde{\coef}_i$ and $\widetilde{\coef}_j$, that reside in a low rank structure reflected by the very similar effective ranks of $\widetilde{\coef}_i$ and $[\widetilde{\coef}_i, \widetilde{\coef}_j]$. For example, in our MNIST experiments using a 4-layer CNN, out of the 6 possible $(i,j)$ pairs, the average effective rank of $\widetilde{\coef}_i$ and $[\widetilde{\coef}_i, \widetilde{\coef}_j]$ are $31.98$ and $38.56$, respectively.
Our observations agrees with and further support recent studies on cross-layer feature correlations \cite{kornblith2019similarity,morcos2018insights,raghu2017svcca}.
Motivated by such empirical observations, we propose to enforce shared dictionary coefficients across layers, and we further extend the sharing to other network sub-structures, e.g., groups of filters within a layer.

\subsection{Coefficients Sharing Across Layers}

Based on the observations above and (\ref{eq:decomposed}), networks with ACDC are constructed by directly sharing $\coef$ within sub-structures as illustrated in Figure~\ref{fig:net}. We introduce two variants named \textit{ACDC-net} and \textit{ACDC-block}.

The simplest variant, \textit{ACDC-net}, is constructed by enforcing common coefficients across all layers in a CNN. 
Formally, given a $N$-layers CNN, the $n$-th convolutional kernel is constructed by
\begin{equation}
\K_n = \atom_n \coef, \forall n = 1, \dots, N.
\end{equation}
Assuming all layers have identical channel number with $c^\prime = c$, the amount of parameters is reduced from $c^2c^2N$ to $c^2m + Nkl^2$.
An illustration of the \textit{ACDC-net} is shown in Figure~\ref{fig:net}. 

\textit{ACDC-block} is a relaxed version of \textit{ACDC-net}. Instead of enforcing common coefficients across all layers, we allow the sharing to happen among a few consecutive layers in a network. We refer a group of consecutive layers with identical number of output channels and identical feature sizes as a \textit{block} in a deep network, and implement \textit{ACDC-block} by enforcing coefficient sharing in each block. 
For example, adopting \textit{ACDC-block} to a VGG16 network \cite{simonyan2014very} is implemented by sharing coefficients within 5 groups, each of which consists of convolutional layers with 64, 128, 256, 512, 512 channels, respectively.

\begin{figure}
	\resizebox{\textwidth}{!}{%

		\begin{minipage}{0.43\linewidth}
			\includegraphics[width=\linewidth]{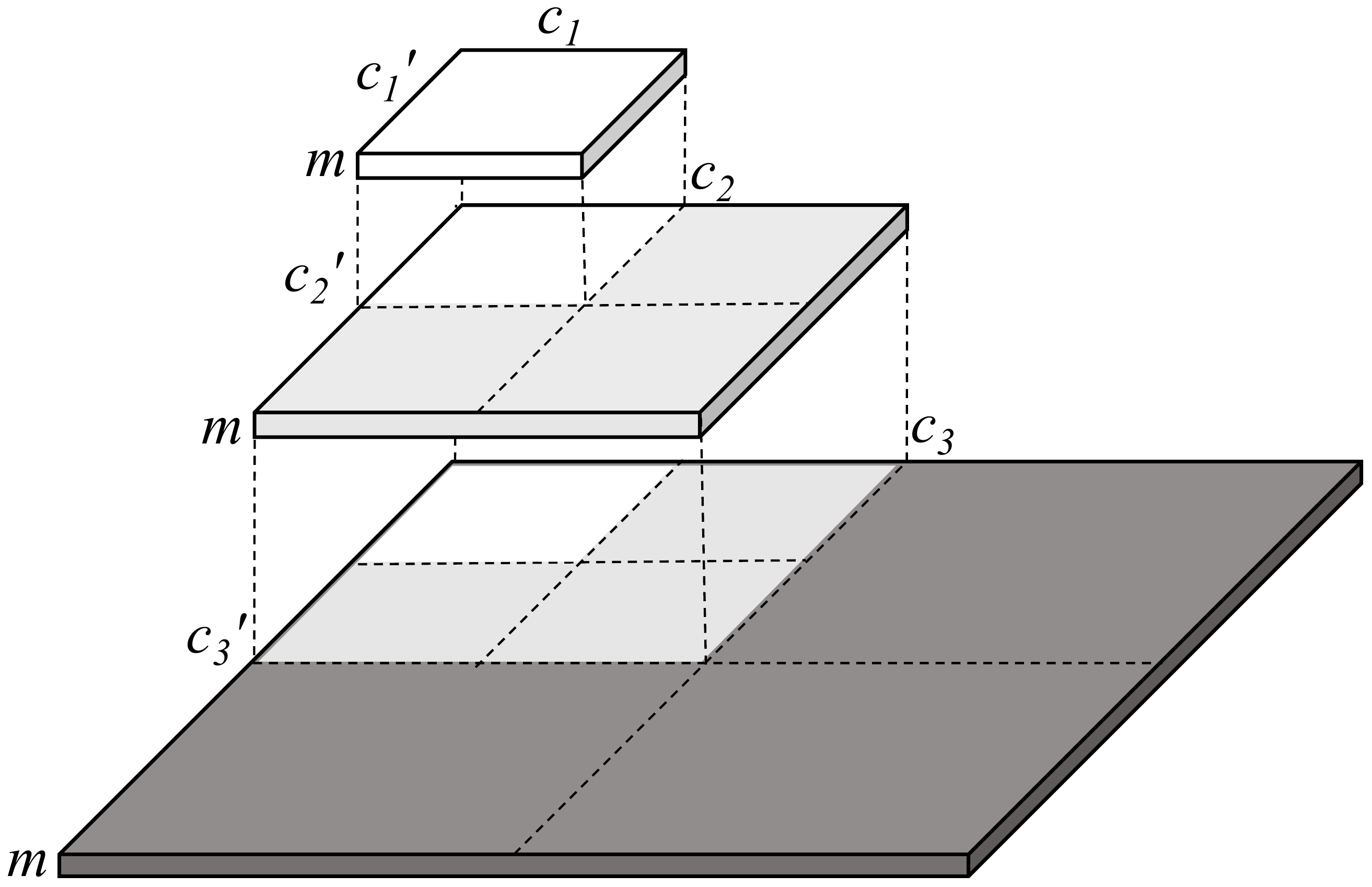}
			\vspace{1mm}
			\caption{Illustration on how coefficients are shared across three layers with increasing numbers of channels. The shared coefficients are initialized with the largest dimensions required.
			}
			\label{fig:index} 
		\end{minipage}
		\hspace{3mm}
		\begin{minipage}{0.7\linewidth}
			\includegraphics[width=\linewidth]{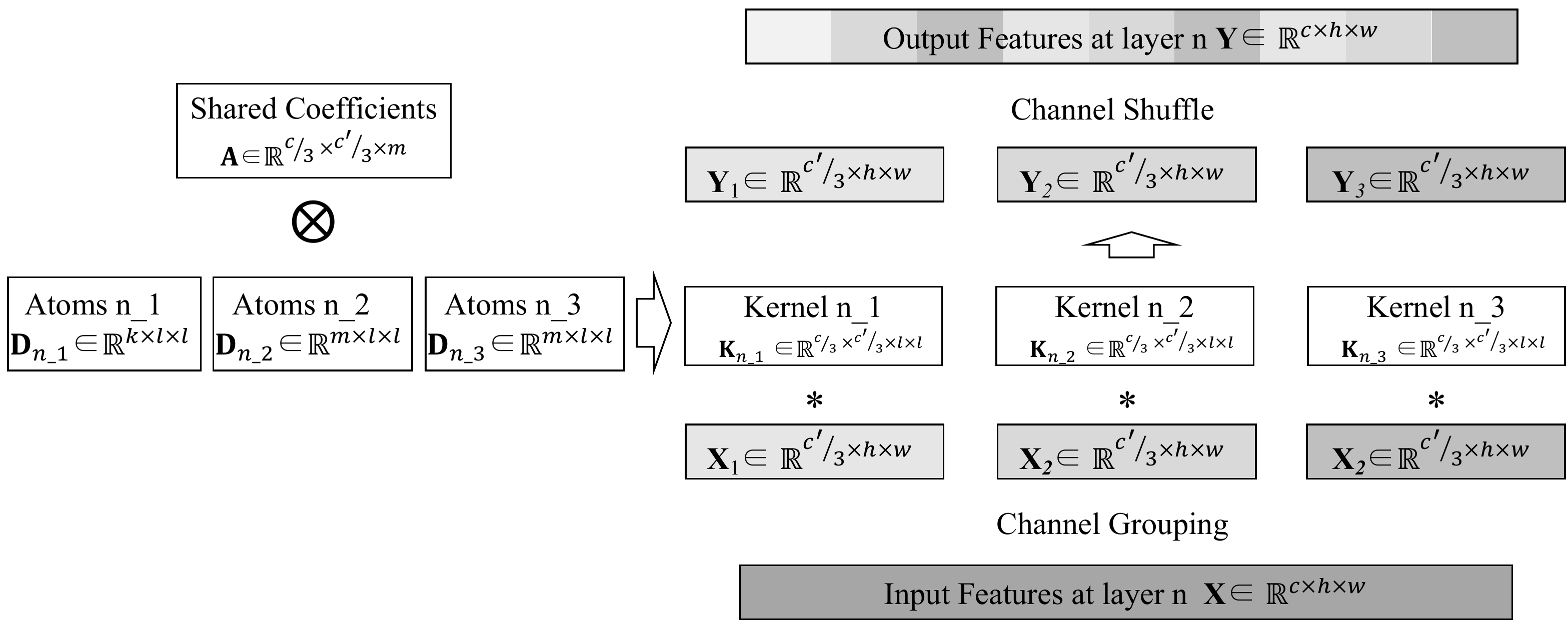}
			\caption{\textit{ACDC with grouping} with three groups at layer $n$. The input feature is first equally divided into groups (denoted as boxes with different grey scales), each of which is convolved with one group of filters reconstructed by multiplying the corresponding filter dictionary atoms and the shared coefficients. The output of three groups are combined by channel shuffle.}
			\label{fig:shuffle} 
		\end{minipage}
	}
\end{figure}

In practice, convolution layers within a network can have different numbers of channels. When sharing coefficients across layers with different channels numbers, we initialize the dimensions of the shared coefficients to be the largest dimensions needed by the corresponding layers. For example, given a $N$-layer CNN with convolutional kernels $\{\mathbf{K}_n \in \mathbb{R}^{c_n \times c_n^\prime \times l \times l}; n = 1, \dots, N\}$, \textit{ACDC-net} is constructed by initializing the shared coefficient as $\coef \in \mathbb{R}^{c_{max} \times c^\prime_{max} \times m}$, where $c_{max} = \max\{c_n; n=1,\dots, N \}$ and $c_{max}^\prime = \max\{c_n^\prime; n=1,\dots, N \}$. The kernels with fewer channels are reconstructed by multiplying the dictionary atoms with a subset of the shared coefficients $\mathbf{K}_n = \atom_n \coef[1:c_n, 1:c_n^\prime, 1:m]$. 
A 3-layer illustration with progressively increased channels is shown in Figure~\ref{fig:index}.
Such a design choice is motivated by multi-scale decomposition \cite{mallat1999wavelet}, and proves to be highly effective with our extensive experimental validation.

\subsection{Coefficients Sharing Across Filter Groups}
While both \textit{ACDC-net} and \textit{ACDC-block} remarkably reduce the number of parameters, the total number of coefficients largely depends on the highest channel number in a deep network. 
For example, a ResNet-18 or a VGG-16 network have up to 512 channels in the last few layers, which become the bottleneck of parameter efficiency. 
Meanwhile, observations in \cite{raghu2017svcca} show that the representation learned at a layer is not fully determined by the number of neurons in the layer, and studies in \cite{michel2019sixteen} reveal the existence of redundancy within a single layer. 
Those observations stimulate us to explore weight sharing within a layer, where redundancy especially in high-dimensional layers can be further squeezed by sharing parameters within groups of filters in a convolutional kernel.

By breaking down the smallest sharing unit from a layer to part of a layer, we propose \textit{ACDC with grouping}, in which a high-dimensional convolutional layer can now be separated into several groups with identical sizes, and sharing coefficient is imposed across groups.
Formally, given a convolutional layer with $c^\prime$ input channels and $c$ output channels, respectively, we divide input channels into $g$ identical-sized groups, and each group is convolved with a convolution kernel $\mathbf{K}_j \in \mathbb{R}^{\frac{c}{g} \times \frac{c^\prime}{g}\times l \times l}, j=1,\dots, g$.
After grouping, we decompose $\{\mathbf{K}_j ; j=1,\dots, g \}$ into shared coefficients $\coef \in \mathbb{R}^{\frac{c}{g} \times \frac{c^\prime}{g}\times m}$, and $g$ independent sets of dictionary atoms $\{ \atom_j \in \mathbb{R}^{m \times l \times l}; j = 1, \dots, g \}$.
In this way, the number of shared coefficients is reduced by $g^2$ times, and the number of dictionary atoms is increased by $g$ times. Since dictionary atoms have orders of magnitude smaller dimension comparing to the coefficients, applying \textit{ACDC with grouping} achieves further parameter reduction. And since each $\K_j$ only convolve with a subset of the input feature, this method reduces the overall computations.

However, directly deploying this sharing mechanism breaks the network into several paralleled subnetworks with no feature passing and gradient propagation among them. 
To remedy this without adding any additional parametric components, we utilize \textit{channel shuffle} \cite{zhang2018shufflenet} that enables information to be efficiently propagated among groups in a non-parametric way. An illustration of the proposed \textit{ACDC with grouping} is presented in Figure~\ref{fig:shuffle}.
Since the size of the shared coefficient now does not depend on the largest feature dimension of a network but the size of the groups, further parameter reduction is achieved by \textit{ACDC with grouping} as shown in Section~\ref{exp}.

\subsection{Regularization by Atom-drop}
To improve the robustness of the dictionary atoms and the corresponding reconstructed kernels, we further propose a structural regularization to dictionary atoms named \textit{atom-drop} inspired by prevalently used dropout \cite{srivastava2014dropout}.
Specifically, when training the network, we randomly drop a dictionary atom with probability $p$, which is referred as \textit{atom drop rate}, by temporarily setting values of the dropped dictionary atom to $0$, and meanwhile values of all other remained dictionary atoms are multiplied by $\frac{1}{1-p}$ in order to maintain a consistent scale of the reconstructed convolutional kernels. At test time, all dictionary atoms are presented with no dropping.

\section{Experiments}
\label{exp}
In this section, we apply variants of ACDC as plug-and-play replacements to the standard convolution, and perform extensive experiments to validate the effectiveness of ACDC as a structural regularization of CNNs. All experiments of ACDC are conducted with the same setting with the underlying model, and atom-drop at all layers with a drop rate of 0.1, if not otherwise specified.
\textit{ACDC with grouping} leads to three variants of ACDC, which are constructed by allowing coefficients to be shared within the entire network, blocks within a network, and layers within a network, and are named as \textit{ACDC-g-net}, \textit{ACDC-g-block}, and \textit{ACDC-g-layer}, respectively.

\subsection{Image Classification}
In this section, we perform standard image classification experiments with the proposed variants of ACDC networks.
\paragraph{Self-comparison on CIFAR-10.}
We first report on CIFAR-10 extensive self-comparison on variants of ACDC constructed with different numbers of dictionary atoms as well as grouping sizes. 
We present performance in terms of both parameter size and classification error in Table~\ref{tab:cifar}.
VGG16 \cite{simonyan2014very}, ResNet18 \cite{he2016deep}, and Wide ResNet (WRN) \cite{zagoruyko2016wide} are adopted as the underlying network architectures in order to show the remarkable compatibility of ACDC.
ACDC enhances deep CNNs with great flexibilities reflected by the wide range of parameter reduction from as low as 98\% reduction with comparable performance, to about 70\% reduction with even higher accuracy.

\begin{table}[]
	\centering 
	\caption{Comparisons on CIFAR-10. The performance on each model are presented by the parameter size and the test error. $m$ and $s$ indicating number of dictionary atoms and grouping size, respectively, e.g., \textit{ACDC-g-net $m8$ $s32$} represents \textit{ACDC-g-net} with 8 dictionary atoms and 32 input and output channels in each filter group. Higher than baseline performances with fewer parameters are marked in bold. Errors are reported by averaged error rates with 5 runs. All networks are trained from scratch.}
	\label{tab:cifar}
	\resizebox{0.8\textwidth}{!}{%
		\begin{tabular}{c|cc|cc|cc|cc}
			\toprule
			\MR{2}*{Architectures} & \MR{2}*{$m$} & \MR{2}*{$s$} & \multicolumn{2}{c|}{VGG16} & \multicolumn{2}{c|}{ResNet18} & \multicolumn{2}{c}{WRN-40-4}  \\
			~ & ~ & ~ & Size & Error & Size & Error & Size & Error \\
			\midrule
			Baseline & - & - &  14.72M & 6.2 & 11.17M & 5.8 & 8.90M & 4.97 \\
			\midrule
			\MR{2}*{\textit{ACDC-net}} &8& - & 2.11M & \textbf{5.67} & 2.28M & 5.9 & 0.58M & \textbf{4.85} \\ 
			~  &16& - & 4.21M & \textbf{5.44} & 4.38M & \textbf{5.4} & 1.11M & \textbf{4.42} \\ 
			\midrule
			\MR{2}*{\textit{ACDC-block}}&8&-& 4.89M & \textbf{5.47} & 2.96M &  \textbf{5.5} & 0.74M& \textbf{4.46} \\ 
			~  &16&-& 9.78M & \textbf{5.40} & 4.38M & \textbf{4.9} & 1.43M & \textbf{4.38}  \\ 
			\midrule
			\MR{2}*{\textit{ACDC-g-net}}&8&32 & 0.03M& 10.24 & 0.20M & 7.3 & 0.07M &  8.20\\ 
			~ &16&64&  0.08M & 9.87 & 0.26M & 7.9 & 0.13M & 6.85   \\ 	
			\midrule
			\MR{2}*{\textit{ACDC-g-block}}&8&32 & 0.06M & 9.71& 0.22M & 5.35 & 0.09M & 8.92\\ 
			~&16&64 & 0.35M & 6.63 & 0.45M & 7.2 & 0.26M & 6.88  \\ 
			\midrule
			\MR{2}*{\textit{ACDC-g-layer}}&8&32 & 0.13M & 6.68 & 0.89M & 5.2 & 0.36M & 5.02\\ 
			~&16&64 & 0.80M & \textbf{5.67} & 0.60M & 6.2  & 1.98M & \textbf{4.23}\\ 
			\bottomrule
		\end{tabular}
	}	
\end{table}

\paragraph{Image classification with comparisons.}
We further present experiments results on CIFAR-10, CIFAR-100, and \textit{Tiny}ImageNet. 
We compare exampled variants of ACDC against HyperNetworks \cite{ha2016hypernetworks} and Soft Parameter Sharing \cite{savarese2019learning}, both of which serve as plug-and-play replacements to standard convolutions as well. 
Though HyperNetworks \cite{ha2016hypernetworks} achieves remarkable parameter reduction, ACDC is able to achieve higher accuracies with even fewer parameters.
The parameter reductions in Soft Parameter Sharing \cite{savarese2019learning} are highly restricted by the large scale elements in the filter bank. For example, SWRN 28-10-1, as the smallest variant of Soft Parameter Sharing on WRN, adopts a single template per sharing group, and can only achieves ~66\% of parameter reduction. By adopting \textit{ACDC-net} and \textit{ACDC-block} to WRN, we are able to achieve both higher parameter reductions and accuracies.
We also compare state-of-the-art light CNN architectures.

\begin{table}[]
	\centering 
	\caption{Classification performances on CIFAR-10, CIFAR-100, and \textit{Tiny}ImageNet datasets. Performance on state-of-the-art light CNN architectures are listed in the upper block. The middle block shows the performance of plug-and-play methods with parameter sharing in CNNs. Performance obtained by our reproductions are marked with $*$. Errors are reported by averaged error rates with 5 runs. All networks are trained from scratch. }
	\label{tab:image}
	\resizebox{\textwidth}{!}{%
		\begin{tabular}{c | c| c c c}
			\toprule 
			Methods & Parameters & CIFAR-10 & CIFAR-100 &  \textit{Tiny}ImageNet \\
			\midrule
			SqueezeNet \cite{iandola2016squeezenet} & 2.36M & 6.98$^*$ & 29.56$^*$ & 48.22$^*$\\
			ShuffleNet \cite{zhang2018shufflenet} & 0.91M & 7.89$^*$ & 29.94$^*$ & 54.72$^*$\\
			ShuffleNet-V2 \cite{ma2018shufflenet} & 1.3M & 8.96$^*$ & 29.68$^*$ &  51.13$^*$\\
			MobileNet-V2 \cite{sandler2018mobilenetv2} &  2.36M  & 5.52$^*$ & 30.02$^*$ & 48.22$^*$\\
			NASNet \cite{zoph2018learning} & 3.1M & 3.59 &  21.77$^*$ & 47.17$^*$ \\ 
			\midrule
			WRN-40-1 HyperNets \cite{ha2016hypernetworks}  & 0.10M & 8.02 & - & - \\ 
			WRN-40-2 HyperNets \cite{ha2016hypernetworks} & 2.24M  & 7.23 & - & - \\ 
			SWRN 28-10-1 \cite{savarese2019learning} & 12M & 4.01 & 19.73 & 43.05$^*$\\
			SWRN 28-10-2 \cite{savarese2019learning} & 17M & 3.75 & 18.37 &  41.12$^*$\\
			\midrule
			WRN-40-1 \textit{ACDC-block $m8$} & 0.043M & 7.19 & 30.23 & 51.47\\ 
			WRN-40-1 \textit{ACDC-block $m24$} & 0.114M & 7.02& 28.14 & 49.05\\
			WRN-40-4 \textit{ACDC-g-layer $m16$ $s32$} & 0.67M & 4.38 &  20.04 & 45.87 \\
			WRN-28-10 \textit{ACDC-g-block $m24$ $s160$} & 2.27M & 4.25 & 19.64 & 41.24\\		
			WRN-28-10 \textit{ACDC-net $m12$ } & 5.21M & 3.52 & 18.81 & 39.96\\	
			WRN-28-10 \textit{ACDC-block $m24$ } & 13.20M & \textbf{3.26} & \textbf{17.85} & \textbf{38.74}\\	
			\bottomrule
		\end{tabular}
		
	}	
\end{table}

\subsection{Adaptation Experiments}
\begin{figure}
	\resizebox{\textwidth}{!}{%
		\begin{minipage}{0.5\linewidth}
			\includegraphics[width=\linewidth]{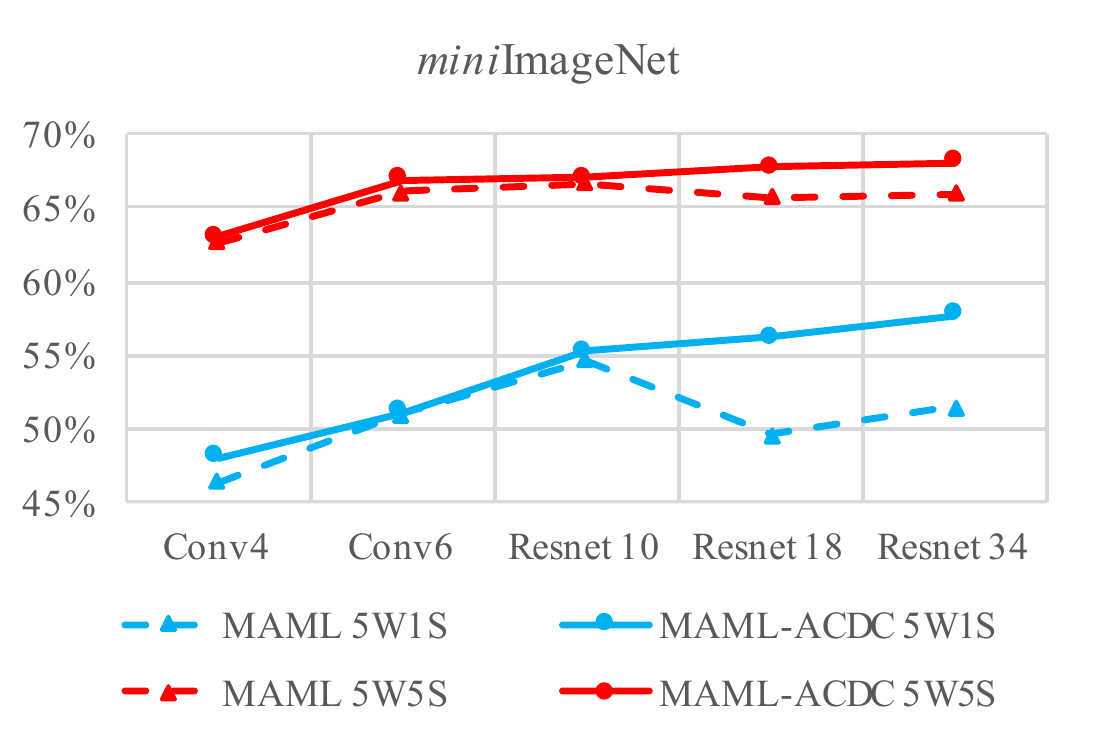}
			\caption{Few-shot image classification with deeper CNN architectures. 5W1S and 5W5S denote 5-way 1-shot and 5-way 5-shot experiments, respectively.}
			\label{fig:maml} 
		\end{minipage}
		\hspace{1mm}
		
		\begin{minipage}{0.55\linewidth}
			\includegraphics[width=\linewidth]{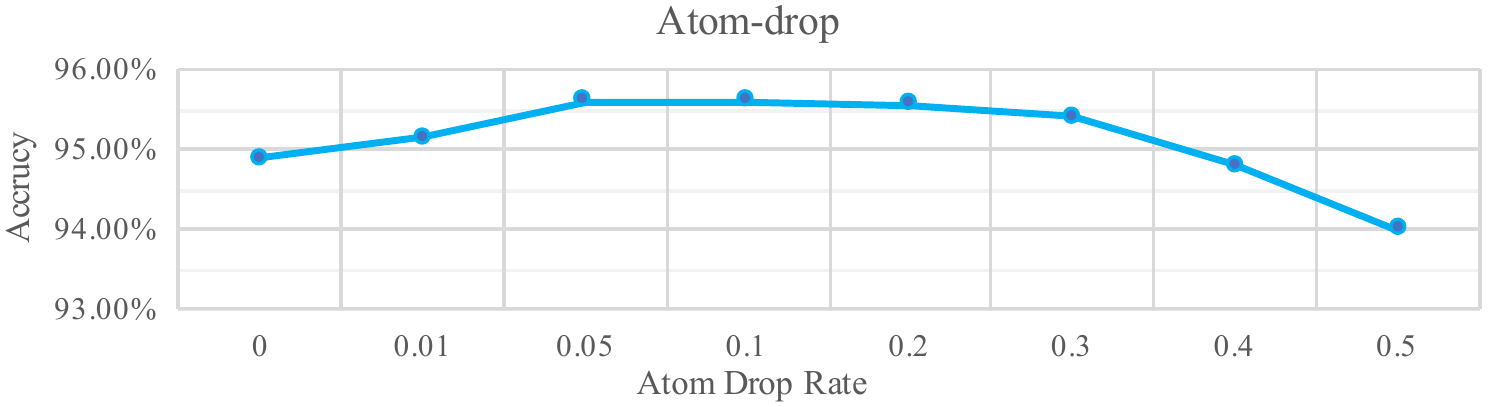}
			\vspace{-5mm}
			\caption{Accuracy with different atom drop rate $p$. 
			}
			\label{fig:atom} 
			\includegraphics[width=\linewidth]{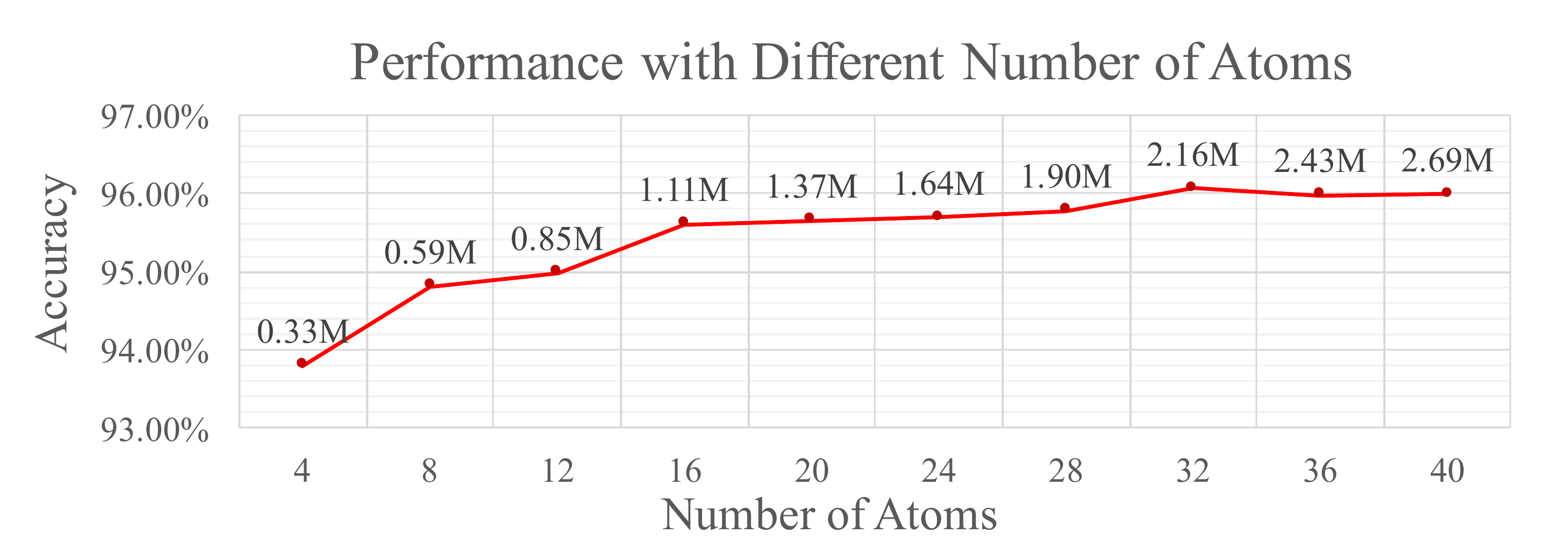}
			\vspace{-5mm}
			\caption{Accuracy with different number of dictionary atoms $m$. Parameter sizes are denoted as $\#M$. 
			}
			\label{fig:basis} 
		\end{minipage}
	}
\end{figure}

We further demonstrate that the proposed ACDC improves the adaptation of deep networks on novel tasks with limited supervisions, which is
validated by few-shot classification using commonly adopted experimental settings. Specifically, we adopt \textit{ACDC-net} on the model-agnostic meta-learning (MAML) \cite{finn2017model} algorithm, which is a method that adapts to a novel task by tuning the entire network from a learned initialization.
Although MAML is designed to be model-agnostic, we consistently observe that it struggles for further performance improvements when using deeper and wider networks. Same observations are  reported in \cite{chen2019closer}. 
We show that such limitation can be alleviated by structural regularizations with ACDC.
We follow the same experimental settings as \cite{chen2019closer} and perform both 5-way 1-shot and 5-way 5-shot image classifications on \textit{mini}ImageNet dataset. The comparisons are shown in Figure~\ref{fig:maml}. 
Though adopting ResNet10 with MAML achieves improvements over simple network with few stacked layers, the performance drops with more residual layers as shown by the results on ResNet18.
By using \textit{ACDC-net} with deeper ResNets, performance is not only maintained but also improved when more layers are used.

\subsection{Computation Efficiency}
ACDC enjoys another merit of being computationally efficient when using \textit{sharing with grouping}. Since after grouping, each group of convolutional filters only convolves with a subset of input features, \textit{ACDC-g-block} and \textit{ACDC-g-net} substantially reduce the number of FLOPs. We report comparisons with ResNet18 and VGG16 in Table~\ref{tab:flops}. All numbers are obtained by feeding the network a typical batch with 100 $64 \times 64$ images. It is clearly shown that by using small groups, the computation can be reduced dramatically, and larger number of dictionary atoms only effects the total FLOPs slightly.

\begin{table}[H]
	\centering 
	\caption{Comparisons of FLOPs with different variants of ACDC with grouping.}
	\label{tab:flops}
	\resizebox{0.9\textwidth}{!}{%
		\begin{tabular}{c | c | c c c c}
			\toprule
			Networks & Baseline & \textit{ACDC m12 s64} & \textit{ACDC m24 s64} & \textit{ACDC m12 s32} & \textit{ACDC m24 s32}\\
			\midrule
			VGG16 & 125.66B & 64.16B & 64.19B & 32.29B & 32.31B \\
			ResNet18 & 222.4B & 116.76B  & 116.80B &  60.13B  & 60.15B\\ 
			\midrule
			
		\end{tabular}
	}	
\end{table}

\subsection{Ablation Studies}
Training with ACDC introduces two hyperparameters, which are the number of dictionary atoms $m$ per sub-structure, and the atom drop rate $p$.
We present here ablation studies on the impacts to the network accuracy with different $p$ and $m$. Both experiments are conducted using ResNet18 with \textit{ACDC-net} and trained on CIFAR-10.
As shown in Figure~\ref{fig:atom}, atom-drop improves generalization when $p \leq 0.1$. Higher values of $p$ potentially degrade the performance as the training becomes unstable. Thus we use $p=0.1$ as the default setting.
As shown in Figure~\ref{fig:basis}, having more dictionary atoms in each sub-structure leads to performance improvements that saturate at $m=32$.
More dictionary atoms also result in larger parameter sizes, which are unfavourable. 

\section{Related Work}
\paragraph{CNN architectures.}
The tremendous success of applying convolutional neural networks (CNNs) on numerous tasks has stimulated rapid developments for more effective and efficient network architectures in both hand-crafted \cite{chen2017dual,he2016deep,howard2017mobilenets,huang2017densely,iandola2016squeezenet,sandler2018mobilenetv2,zhang2018shufflenet} and automatically discovered \cite{elsken2018neural,liu2018progressive,pham2018efficient,zoph2016neural} manners.
We consider our work orthogonal to such topology-based methods, as the plug-and-play property of the proposed ACDC allows it to be equipped to all the aforementioned methods as a replacement to the standard convolution.
Besides efforts on studying efficient network architectures, methods for network compression and pruning \cite{han2015deep,han2016dsd,han2015learning,he2017channel,li2016pruning,luo2017thinet} have been extensively studied for decreasing the model size by pruning the inconsequential connections and weights of a network.
Methods \cite{ha2016hypernetworks,savarese2019learning} align with our direction as they are also insensitive to network topologies. And as shown in the experiments, ACDC can achieves higher performance in terms of parameter reduction and classification accuracy with greater flexibility.

\paragraph{Kernel decomposition in CNNs}
Convolutional kernel decomposition has been studied for various objectives. 
\cite{sosnovik2019scale} utilizes kernel decomposition as a tool of allowing same kernel with multiple reception field to be constructed without interpolations.
DCFNet \cite{qiu2018dcfnet} is proposed as a principle way of regularizing the convolutional filter structures by decomposing convolutional filters in CNN as a truncated expansion with pre-fixed bases.

\begin{figure}[h]
	\centering
	\includegraphics[width=\linewidth]{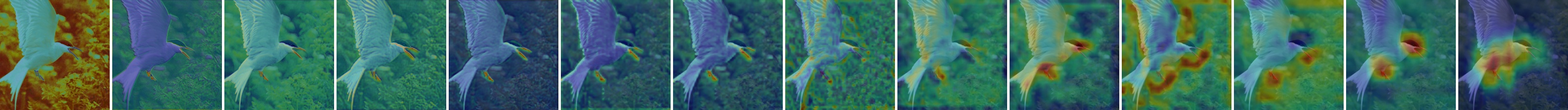}\\
	\vspace{0.5mm}
	\includegraphics[width=\linewidth]{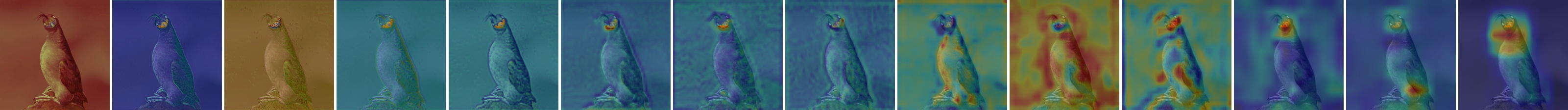}\\
	\vspace{0.5mm}
	\includegraphics[width=\linewidth]{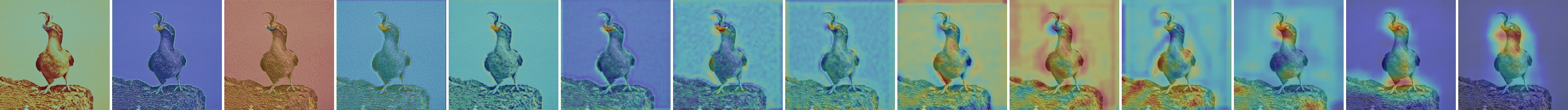}\\
	\vspace{0.5mm}
	\includegraphics[width=\linewidth]{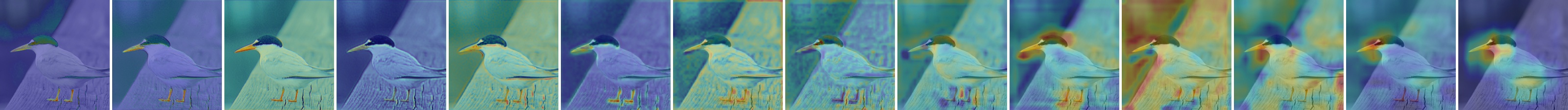}\\
	\caption{Illustration on extending CAM from the last layer to all layers with \textit{ACDC-net}. In CAM, the visualized heatmap explains the importance of image regions. Each row shows the class activation maps of a sample from the first convolution layer (left) to the final convolution layer (right).
	}
	\label{fig:cam} 
\end{figure}

\section{Conclusion and Interpretability Discussion}
In this paper, we introduced atom-coefficient decomposed convolution, a plug-and-play replacement to the standard convolution by imposing structural regularization to kernels in a CNN.
We presented observations that, due to the underlying cross-layer correlations, coefficients in the decomposed convolution layers reside in a low rank structure.
We explicitly exploited such observations by enforcing coefficients to be shared within sub-structures of CNNs, and achieved significant parameter reductions. Variants of ACDC can be constructed with different sharing structures, number of atoms, and grouping sizes. We reported extensive experiment results showing the effectiveness of ACDC on standard image classification and adaptations. 

The structural regularization with ACDC has the potential for better interpretability of CNNs, due to the cross-layer shared coefficients.
We close our paper with an illustration that extends class activation mapping (CAM) \cite{zhou2016learning}, which is originally proposed to explain the importance of image regions but only at the final convolution layer. CAM is calculated by weighted averaging the features of the final convolution layer by the weight vector of a particular class. 
Since now with \textit{ACDC-net}, feature across layers are generated with the same coefficients, we exploit the potential correspondence to extend CAM to all the preceding layers using the same weighted sum, and visualize the activation maps for all layers as in Figure~\ref{fig:cam}. 
We use a VGG16 network with 13 convolution layers and train it on CUB-200 \cite{wah2011caltech} high-resolution bird classification dataset. 
It is clearly shown that, while the activation maps for shallow layers are inevitably noisy due to the limited reception fields, features in deeper layers are progressively refined to the discriminative regions.
This shows the great potential for better interpretability with ACDC, and we will keep this as a direction of future effort.


\bibliographystyle{plain}
\bibliography{egbib}

\end{document}